\documentclass[smallcondensed]{svjour3}
\usepackage{graphicx}
\usepackage{color}
\usepackage{url}
\usepackage{theapa}

\journalname{Artificial Intelligence Review}

\begin{document}

\title{
Software Verification and Graph Similarity for Automated Evaluation of
Students' Assignments\thanks{This work was partially supported by the Serbian Ministry of Science grant
174021 and by Swiss National Science Foundation grant SCOPES IZ73Z0\_127979/1.}}

\titlerunning{Verification and Similarity for Automated Evaluation of  Students' Assignments}

\author{Milena Vujo\v{s}evi\'c-Jani\v{c}i\'c \and Mladen Nikoli\'c \and Du\v{s}an To\v{s}i\'c \and Viktor Kuncak}
\institute{Milena Vujo\v{s}evi\'c-Jani\v{c}i\'c \and Mladen Nikoli\'c \and Du\v{s}an To\v{s}i\'c \at
            Faculty of Mathematics, University of Belgrade, Belgrade, Serbia\\
              \email{milena@matf.bg.ac.rs}\\
              \email{nikolic@matf.bg.ac.rs}\\
              \email{dtosic@matf.bg.ac.rs}\\
           \and
           Viktor Kuncak,
          School of Computer and Communication Sciences, EPFL, Station 14, CH-1015 Lausanne, Switzerland\\
              \email{viktor.kuncak@epfl.ch}\\
}

\date{Received: date / Accepted: date}

\maketitle

\begin{abstract}
In this paper we promote introducing software
verification and control flow graph
similarity measurement in automated evaluation of students' programs.
We present a new grading framework that merges results obtained by
combination of these two approaches with results obtained by
automated testing, leading to improved quality and precision of
automated grading. These two approaches are also useful in providing
a comprehensible feedback that can help students to improve the
quality of their programs
We also present our corresponding tools
that are publicly available and open source. The tools are based on
LLVM low-level intermediate code representation, so they could be
applied to a number of programming languages. Experimental evaluation
of the proposed grading framework is performed on a corpus of
university students' programs written in programming language C.
Results of the experiments show that automatically generated grades
are highly correlated with manually determined grades suggesting that
the presented tools can find real-world applications in studying and
grading.

\end{abstract}

\section{Introduction}
\label{sec:intro}

Automated evaluation of programs is beneficial for both teachers and
students \cite{Pears07}. For teachers, automated evaluation is helpful
in grading assignments and it leaves more time for other activities with students. For students, it provides immediate feedback
which is very important in process of studying, especially in computer science where students take a challenge of making the computer follow their  intentions \cite{nipkow}. Immediate feedback is particularly helpful at first programming courses where students have frequent and deep misconceptions \cite{mvjtosic}.

Benefits of automated evaluation of programs are even more significant in the context of online learning. A number of world's leading universities
offer numerous online courses. The number of students taking such courses
is measured in millions and quickly growing \cite{Allen09}.
In online courses, the teaching process is carried out on the computer,
the contact with teacher is already minimal and hence the fast and
substantial automatic feedback is especially desirable. Therefore,
automation of evaluation tasks in online learning is very important.

Most of the tools for automated evaluation of students' code are
based on automated testing \cite{douce05}. Testing is used for
checking functional correctness of student's solution, i.e., whether
the student's program exhibits the desired behavior on selected
inputs. Testing can also be used for
detecting bugs. We consider bugs to be runtime errors and exclude errors that only
compromise functional correctness (for example, in programming
language C, some important bugs are buffer overflow, null pointer
dereferencing and division by zero). Although there is a variety of
software verification tools that could enhance automated bug finding
in students' programs (by analyzing the code without executing it),
these tools are usually too complex to use and cannot be easily adapted for educational purposes.

In addition to checking functional correctness, an evaluation tool
may also analyze program efficiency and/or complexity by profiling.
Relevant aspects of program quality are also it's design and
modularity (adequate decomposition of code to functions). These
issues are addressed by checking similarity to a teacher provided solution.
In order to check similarity, aspects that can be analyzed are: frequencies of
keywords, number of lines of code, number of variables etc. Recently, a more sophisticated approach of grading students' programs
by measuring the similarity of related graphs
has been proposed \cite{Wang07,Naude10}.
Recent surveys of related approaches are given elsewhere \cite{mutka05,ihantola10}.

In this paper, we propose a new grading framework for automated
evaluation of students' programs aiming primarily at introductory
programming courses. The framework is based on merging information
from three different evaluation methods: it merges results obtained by
software verification (automated bug finding) and control flow graph
(CFG) similarity measurement with results obtained by automated
testing. The synergy between automated testing, verification, and
similarity measurement improves the quality and precision of
automated grading and overcoming the individual weaknesses of these approaches. Our experimental results show that our framework can
lead to a grading model that highly correlates to manual grading and
therefore gives promises for real-world applicability in education.

We also briefly discuss tools for software verification \cite{vstteLAV12} and CFG similarity \cite{Nikolic13}, that we use for assignment evaluation. These tools, based on novel methods, are publicly available and open
source.\footnote{\url{http://argo.matf.bg.ac.rs/?content=lav}}
Both tools use the low-level intermediate code representation
LLVM. Therefore, they could be applied to a number of programming languages and
could be complemented with other existing LLVM based tools (e.g.,
tools for automated test generation).  Also, the tools are enhanced
with support for meaningful and comprehensible feedback to students,
so they can be used both in the process of studying and in the
process of grading assignments.

\paragraph{Overview of the paper.}
Necessary background information is given in Section \ref{sec:background}. Motivating
examples for the synergy of the three proposed approaches are given
in Section \ref{sec:synergy}. The grading setting and the corpus
used for evaluation are described in Section \ref{sec:grading}.
The role of the verification techniques in automated
evaluation is discussed in
Section \ref{sec:verification} and the role of structural similarity
measurement is discussed in Section \ref{sec:similarity}. An
experimental evaluation of the proposed framework for automated
grading is presented in Section \ref{sec:ag}. Section \ref{sec:rw}
contains information about related work.
Conclusions and outlines of possible directions of future work are given in Section \ref{sec:conclusions}.

\section{Background}
\label{sec:background}
This section provides an overview of intermediate languages, the LLVM
tool, software verification, the LAV tool, control flow graphs and
graph similarity measurement.

\paragraph{Intermediate languages and LLVM.}
An intermediate language separates concepts and
semantics of a high level programming language from low level issues
relevant for a specific machine. Examples of intermediate languages
include the ones used in LLVM and .NET framework.
LLVM\footnote{\url{http://llvm.org/}} is an open source, widely used,
rich compiler framework, well suited for developing new
mid-level language-independent analyses
and optimizations of all
sorts \cite{llvm}. LLVM intermediate language is assembly-like
language  with simple RISC-like instructions. It provides easy
construction of control flow graphs of program functions and of
entire programs. There is a number of tools using LLVM for various
purposes, including software verification. LLVM has front-ends for C,
C++, Ada and Fortran, while there are external
projects for translating a number of other languages to LLVM intermediate
representation (e.g., Python, Ruby, Haskell, Java, D, PHP, Pure, and
Lua).

\paragraph{Software verification and LAV.}
Verification of software and automated bug finding are some of the
greatest challenges in computer science. Software bugs cost the world
economy billions of dollars annually \cite{tassey02economic}.
Software verification tools aim at automatically checking correctness
properties. Different approaches to automated checking of
software properties exist, such as symbolic execution \cite{King76}, model
checking \cite{modelc} and abstract interpretation \cite{cousot77}.
Software verification tools usually use automated theorem provers.

LAV \cite{vstteLAV12} is an open-source tool for statically
verifying
program assertions and locating bugs such as buffer overflows, pointer
errors and division by zero. LAV uses popular LLVM
infrastructure. As a result, it supports several programming
languages that compile into LLVM, and benefits from the robust LLVM
front ends. LAV is primarily aimed at programs in the C programming
language, in which the opportunities for errors are abundant. For each
safety critical command, LAV generates a first order logic formula
that represents its correctness condition. This formula is checked by
one of the several SMT solvers \cite{BSST09HBSAT} used by LAV.
If a command cannot be proved to be safe, LAV translates a potential
counterexample from the solver into a program trace that exhibits
this error. It also extracts the values of relevant program variables
along this trace. LAV was already used, to a limited extent,
for automated bug finding in students' assignments \cite{vstteLAV12}.

\paragraph{Control flow graph.}
A control flow graph (CFG) is a graph-based representation of all
paths that might be traversed through a program during its execution.
Each node of CFG represents a sequence of commands containing only
one path of execution (there are no jumps, loops, conditional
statements, etc.). The control flow graphs can be produced by various
tools, including LLVM. A control flow  graph clearly separates the
structure of the program and its contents. Therefore, it is a suitable
representation for structural comparison of programs.

\paragraph{Graph similarity and neighbor matching method.}
There are many similarity measures for graphs and their nodes
\cite{Kleinberg99,Heymans03,Blondel04,Nikolic13}. These measures have
been successfully applied in several practical domains like ranking
of query results, synonym extraction, database structure matching,
construction of phylogenetic trees, analysis of social networks, etc.
A short overview of similarity measures for graphs can be found in the literature
\cite{Nikolic13}.

A specific similarity measure for graph nodes called \emph{neighbor matching}, possesses properties relevant for our purpose
that other similar measures lack \cite{Nikolic13}. It allows
similarity measure for graphs to be defined based on similarity
scores of their nodes. The notion of similarity of nodes is based on
the intuition that \emph{two nodes $i$ and $j$ of graphs $A$ and $B$
are considered to be similar if neighbor nodes of $i$ can be matched
to similar neighbor nodes of $j$}. More detailed definitions
follow.

In the neighbor matching method, if a graph contains an edge $(i,j)$,
the node $i$ is called an {\em in-neighbor} of node $j$ in the graph
and the node $j$ is called an {\em out-neighbor}  of the node $i$ in
the graph. An {\em in-degree} $id(i)$ of the node $i$ is the number
of in-neighbors of $i$, and an {\em out-degree} $od(i)$ of the node
$i$ is the number of out-neighbors of $i$.

If $A$ and $B$ are two finite sets of arbitrary elements, a
\emph{matching} of elements of sets $A$ and $B$ is a set of pairs
$M=\{(i,j)|i\in A,j\in B\}$ such that no element of one set is paired
with more than one element of the other set. For the matching  $M$,
\emph{enumeration functions} $f:\{1,2,\ldots k\}\rightarrow A$ and
$g:\{1,2,\ldots k\}\rightarrow B$ are defined such that
$M=\{(f(l),g(l))|l=1,2,\ldots,k\}$ where $k=|M|$. If $w(a,b)$ is a
function assigning weights to pairs of elements $a\in A$ and $b\in
B$, the \emph{weight of a matching} is the sum of weights assigned to
the pairs of elements from the matching. The goal of the
\emph{assignment problem} is to find a matching of elements of $A$
and $B$ of the highest weight (if two sets are of different
cardinalities, some elements of the larger set will not have
corresponding elements in the smaller set). The assignment problem is
usually solved by the well-known Hungarian algorithm of complexity
$O(mn^2)$ where $m=\max(|A|,|B|)$ and $n=\min(|A|,|B|)$
\cite{Kuhn55}, but there are also more efficient algorithms.

The calculation of similarity of nodes $i$ and $j$, denoted $x_{ij}$,
is based on  iterative procedure given by the following equations:
$$x^{k+1}_{ij} \leftarrow \frac{s^{k+1}_{in}(i,j)+s^{k+1}_{out}(i,j)}{2}$$
where
\begin{equation}
s^{k+1}_{in}(i,j) \leftarrow \frac{1}{m_{in}}\sum_{l=1}^{n_{in}}x^k_{f^{in}_{ij}(l)g^{in}_{ij}(l)}\hspace{1cm}s^{k+1}_{out}(i,j) \leftarrow \frac{1}{m_{out}}\sum_{l=1}^{n_{out}}x^k_{f^{out}_{ij}(l)g^{out}_{ij}(l)}
\label{eq:updaterule}
\end{equation}
$$m_{in}=\max(id(i),id(j))\hspace{1cm}m_{out}=\max(od(i),od(j))$$
$$n_{in}=\min(id(i),id(j))\hspace{1cm}n_{out}=\min(od(i),od(j))$$
where functions $f^{in}_{ij}$ and $g^{in}_{ij}$ are the enumeration
functions of the optimal matching of in-neighbors for nodes $i$ and
$j$ with weight function $w(a,b)=x^k_{ab}$, and analogously for
$f^{out}_{ij}$ and $g^{out}_{ij}$. In Equations \ref{eq:updaterule},
$\frac{0}{0}$  is defined to be 1 (used in case when
$m_{in}=n_{in}=0$ or $m_{out}=n_{out}=0$). Initial similarity values
$x^0_{ij}$ are set to 1 for each $i$ and $j$.
The termination condition is $\max_{ij}
|x^{k}_{ij}-x^{k-1}_{ij}| < \varepsilon$ for some chosen precision
$\varepsilon$ and the iterative algorithm is proved to converge
\cite{Nikolic13}.

The similarity matrix $[x_{ij}]$ reflects the similarities of nodes of
two graphs $A$ and $B$. The  similarity of the graphs can be defined
as the weight of the optimal matching of nodes from $A$ and $B$
divided by the number of matched nodes \cite{Nikolic13}.

\section{The Need for Synergy of Testing, Verification, and Similarity Measurement}
\label{sec:synergy}

Automated testing of programs is a very important part of the evaluation
process. Unfortunately, the grading system is directly influenced by the
choice of test cases. Also, no matter
whether the test cases are automatically generated or manually designed, testing cannot
guarantee neither functional correctness of a program nor the absence of bugs.

For checking functional correctness, combination of random testing
with  eval\-uator-supplied test cases is a common choice
\cite{MandalMR07}. However, randomly generated test cases are not
likely to hit a bug if it exists \cite{godefroid12}, while manually
choosing all important test cases is not a trivial job and can be time consuming. It is not sufficient that test cases cover all
important paths through the program. It is also important to carefully
choose values of the variables for each path --- for some values
along the same path a bug can be detected while for some other
values the bug can stay undetected.

Also, manually generated test cases are
designed according to the expected solutions, while the
evaluator cannot predict all the important paths through the
student's solution. Even running a test case that hits a certain bug
(for example, a buffer overflow bug in a C program) does not
necessarily lead to any visible undesired behavior if the running is
done in a normal (or sandbox) environment. Finally, if one manages to hit
a bug by a test case, if the bug produces the {\em Segmentation
fault} message, it is not a feedback that student can easily
understand and use for debugging the program. In the context of
automated grading, this feedback cannot be easily used since it may
have different causes. In contrast to program testing, software
verification tools like Pex \cite{PEX}, Klee \cite{KLEE}, S2E
\cite{S2E}, CBMC \cite{CBMC}, ESBMC \cite{ESBMC}, and LAV
\cite{vstteLAV12} can give much better explanations (e.g., the kind of
bug and the program trace that introduces an error).

\begin{figure}[h!]
\begin{verbatim}
0:  #define max_size 50
1:  void matrix_maximum(int a[][max_size], int rows, int columns, int b[])
2:  {
3:      int i, j, max=a[0][0];                  int i, j, max;
4:      for(i=0; i<rows; i++)                   for(i=0; i<rows; i++)
5:      {                                       {
6:                                                  max = a[i][0];
7:          for(j=0; j<columns; j++)                for(j=0; j<columns; j++)
8:              if(max < a[i][j])                       if(max < a[i][j])
9:                 max = a[i][j];                          max = a[i][j];
10:         b[i] = max;                             b[i] = max;
11:         max=a[i+1][0];
12:     }                                       }
13:     return;                                 return;
14:  }
\end{verbatim}
\caption{Buffer overflow in the code on left-hand side cannot be
discovered  by simple testing. Functionally equivalent solution
without a bug is given on right-hand side.}
\label{fig:c1}
\end{figure}

The example function shown at Figure \ref{fig:c1} is extracted from a
student's code written on an exam.
It calculates the maximum value of
each row of a matrix and writes these values into an array. This
function is used in a context where the memory for the matrix is
statically allocated and  numbers of rows and columns are less or
equal to the allocated sizes of the matrix. However, in the line
 $11$, there is a possible buffer overflow bug, since $i+1$ can
exceed the allocated number of rows for the matrix. It is possible
that this kind of a bug does not affect the output of the program or
destroy any data, but  in a slightly different context it
can be  harmful, so students should be warned and penalized for
making such errors. The bugs like this one can be missed in testing but are
easily discovered by verification tools like LAV.

Functional correctness and absence of bugs are not the only important
aspects  of students' programs. The programs are often supposed to
meet certain requirements concerning the structure of the program,
such as its modularity (adequate decomposition of code to functions) or
simplicity. Figure \ref{fig:similarity} shows two solutions of
different modularity or structural simplicity for two problems.
Neither testing, nor software verification can be used to assess
these aspects of the programs. This problem can be addressed by
checking the similarity of student's solution with a teacher provided solution,
i.e., by analyzing the similarity of their related graphs (e.g. CFGs)
\cite{Wang07,Naude10,Nikolic13}.\footnote{In Figure 2, the second example could also be distinguished by profiling for large inputs,
because it is quadratic in one case and linear in the other. However, profiling cannot be used to assess structural
properties in general.}

\begin{figure}[h!]
\begin{center}
\begin{tabular}{|c|l|l|}
\hline
Problem &First solution & Second solution\\
&&\\
\hline \hline
&&\\
&{\texttt{if(a<b) n = a;}}  & {\texttt{n = min(a, b);}} \\
&{\texttt{else n = b;}} & \\
1.&{\texttt{if(c<d) m = c;}} & {\texttt{m = min(c, d);}} \\
&{\texttt{else m = d;}} \verb|     | &\\
&& \\
\hline
&& \\
&{\texttt{for(i=0; i<n; i++)}}&                 {\texttt{for(i=0; i<n; i++)}} \\
&\verb|     | {\texttt{for(j=0; j<n; j++)}} &               \verb|     |  {\texttt{m[i][i] = 1;}} \\
2. &\verb|     | \verb|     | {\texttt{if(i==j)}} & \\
&\verb|     | \verb|     | \verb|     | {\texttt{m[i][j] = 1;}} & \\
&& \\ \hline

\end{tabular}
\end{center}
\caption{Examples extracted from two students' solutions of the same problem}
\label{fig:similarity}
\end{figure}

Finally, using similarity only (like in \cite{Wang07,Naude10}) or
even with support of a bug finding tool, would miss to penalize incorrectness of
program's behavior. Figure \ref{fig:testing} gives a simple example program, extracted from a real student's solution, that is
very similar to the expected solution and without verification errors. However, this program is
not functionally correct. Therefore, we conclude that the synergy of these three approaches is needed for
sophisticated evaluation of students' assignments.

\begin{figure}[h]
\begin{center}
\begin{verbatim}
    max = 0;                    max = a[0];
    for(i=0; i<n; i++)          for(i=1; i<n; i++)
        if(a[i] > max)              if(a[i] > max)
            max = a[i];                 max = a[i];
\end{verbatim}
\end{center}
\caption{Code extracted from student's solution (left-hand side) and
expected solution (right-hand side).  In the student's solution there
are no verification bugs, it is very similar to the expected solution
but it does not perform the desired behavior (in the case when all
elements of the array \texttt{a} are negative integers).}
\label{fig:testing}
\end{figure}

\section{Grading Setting}
\label{sec:grading}
There may be different grading settings depending on aims of the
course and goals of teachers.  The setting used at an introductory
course of programming in C (at University of Belgrade) is
taking
exams on computers and expecting from students to write working
programs. In order to help students achieve this goal, each assignment is
 provided with several test cases which illustrate
desired behavior of the solution. Students are also provided with
sufficient (but limited) time for developing and testing programs. If
a student fails to provide a working program that gives correct
results for given test cases, his/her solution is not further
examined. Otherwise, the program is tested by additional test
cases (unknown to students) and a certain amount of points is given corresponding to the
test cases successfully passed. Only if all these test cases are
successfully passed, the program is further manually examined and may
obtain additional points with respect to other features of the
program (efficiency, modularity, simplicity, absence of bugs, etc).

All experiments described in this paper were preformed on a corpus of
programs written by students on the exams, following the
described grading setting. The corpus consists of 266 solutions to 15
different problems. These problems include numerical calculations,
manipulations with arrays and matrices, manipulations with strings,
and manipulations with data structures. Only programs that passed all
test cases were included in this corpus. These programs are the main
target of our automated evaluation technique since the manual grading
was applied only in this case and we want to explore potentials for
completely eliminating manual grading. These programs obtained 80\%
of the maximal score (as they passed all test cases) and additional
potential 20\% were given by manual inspection. The grades are
expressed at the scale from 0 to 10. The corpus together with problem descriptions and the final marks are
publicly available.\footnote{\url{http://argo.matf.bg.ac.rs/?content=lav}}

\section{Assignment Evaluation and Software Verification}
\label{sec:verification}

In this section we show benefits of using software verification tool
in assignment evaluation, e.g., generating useful feedback for
students and providing improved assignment evaluation for teachers.

\subsection{Software verification for assignment evaluation}

No software verification tool can report all the bugs in a program
without introducing false alarms  (due to the undecidability of the
halting problem). False alarms (i.e., reported "bugs" that are not
real bugs) arise as a consequence of approximations that are
necessary in modeling of programs.

The most important approximation is concerned with dealing with
loops. Different verification approaches  use various techniques for
dealing with loops. These techniques range from under-approximations
of loops to over-approxima\-tions of loops. Under-approximation of
loops, as in bounded model checking techniques \cite{modelc}, uses a
fixed number $n$ for loop unwinding. In this case, if the code is
verified successfully, it means that the original code has no bugs
for $n$ or less passes through the loop. However, it may happen that
some bug remains undiscovered if the unwinding is performed an insufficient number of times. Over-approximation of loops can be done
by simulation of first $n$ and last $m$ passes through the loop \cite{vstteLAV12} or by
using abstract interpretation techniques \cite{cousot77}. If there
are no bugs detected in the over-approximated code, then the original
code has no bugs too. However, in this case, a false alarm can appear
after or inside a loop. On the other hand, precise dealing with
loops, like in symbolic execution techniques, can be non terminating.

False alarms are highly unwelcome in software development, but still
are not critical --- the developer can fix the problem or confirm
that the reported problem is not really a bug (and both of these are
situations that the developer can expect and understand). However,
false alarms in assignment evaluation are rather critical and have to
be eliminated. For teachers, there should be no false alarms, because
the evaluation process should be as automatic and reliable as
possible. For students, there should be no false alarms because they
would be confused if told that something is a bug when it is not. In
order to eliminate false alarms, a system may be non-terminating or
may miss to report some real bugs. In assignment evaluation, the
second choice is more reasonable --- the tool has to be terminating,
must not introduce false alarms, even if the price is missing some
real bugs. These requirements make applications of software
verification in education rather specific, and special care has to be
taken when these techniques are applied.

\subsection{LAV for assignment evaluation}
\label{subsec:lavassign}

LAV is a general purpose verification tool and has a number of
options that can adapt its behavior to the desired context. When running LAV in the assignment evaluation context,
most of these options can be fixed.

The most important choice for the user is the choice of the way in
which LAV deals with loops. LAV has support for both
over-approximation of loops and for fixed number of unwinding of
loops (under-approximation), two common techniques for dealing with loops.
Setting up the upper loop bound (if under-approximation is used),
is problem dependent and should be done by the teacher for each assignment.

We use LAV in the following way. LAV is first invoked with its default
parameters --- over-approximation of loops. Since this technique can
introduce false alarms, if a potential bug is found after or inside a
loop, the verification is invoked again but this time with fixed
unwinding parameter. If the bug is still present, then it is
reported. Otherwise, the previously detected potential bug is
considered to be a false alarm and it is not reported.

In software verification, each detected bug is important and
should be reported. However, some bugs can confuse novice
programmers, like the one shown in Figure \ref{ex:malloc}. In this
code, at the line  \texttt{11}, there is a possible buffer
overflow. For instance, for $n = 0x80000001$ only $4$ bytes will be allocated for the pointer
\texttt{array}, because of an integer
overflow. This is a verification error, but a teacher may
decide not to consider this kind of bugs. For this purpose, LAV can
be invoked in mode for students (so the bugs like this one are not reported).

\begin{figure}
\begin{verbatim}
    1:  unsigned i, n;
    2:  unsigned *arr;
    3:  scanf("%u", &n);
    4:  array = malloc(n*sizeof(unsigned));
    5:  if(array == NULL)
    6:  {
    7:      fprintf(stderr, "Unsuccessful allocation\n");
    8:      exit(EXIT_FAILURE);
    9:  }
    10: for(i=0; i<n; i++)
    11:    array[i] = i;
\end{verbatim}
\caption{Buffer overflow in this code is a verification error, but
the teacher may decide not to consider  this kind of bugs.}
\label{ex:malloc}
\end{figure}

To a limited extent, LAV was already used on students' assignments
at an introductory programming course \cite{vstteLAV12}. In these
experiments, most of the programs from the corpus were not
functionally correct. It was shown that the vast majority of bugs, produced by students,
follow wrong expectations --- for instance,
expectations that input parameters of their programs will meet
certain constraints and that memory allocation will always succeed.
It is also noticed that most of the reported bugs are consequence of
only few oversights. In many cases, omission of a necessary check
produces several bugs in the rest of the program. Therefore,
the number of bugs, as reported by a
verification tool, is not a reliable indicator of program
quality. This property will be taken into account in automated grading.

\subsection{Experimental evaluation}

As discussed in Section \ref{sec:synergy}, programs that successfully pass a testing phase can still contain bugs. To show that this problem is
practically important, we used LAV to analyze programs from the
corpus described in Section \ref{sec:grading}.

For each problem, LAV was ran with its default parameters, and programs with potential bugs were checked with under-approximation of loops,
as described in Section \ref{subsec:lavassign}.\footnote{When analyzing the solutions of problems 3, 5 and 8, only under-approximation of loops was used. This was the consequence of the formulation of
the problems given to the students. Namely, the formulation of these
problems contained some assumptions on input parameters. These
assumptions implied that some potential bugs should not be considered
(because these are not bugs when these additional assumptions are
taken into account).} The results are shown
in Table \ref{tab:rezultati}. The time that LAV spent in analyzing
the programs was typically negligible.\footnote{Generally, in this context, a time limit can be given to the verification tool and if it was exceeded no bug will be reported (in order to avoid reporting false alarms) or a program can be checked using the same parameters but with another underlying solver (if applicable for the tool).}
LAV discovered bugs in 35 solutions that
successfully passed the testing.
There was one bug missed by manual
inspection and detected by LAV and one bug missed by LAV and detected by manual
inspection. The bug missed by manual inspection was the one described in Section \ref{sec:synergy} and given in Figure \ref{fig:c1}. The bug missed by LAV was a consequence of the problem formulation which was too general to allow a precise unique upper loop unwinding parameter value for all possible solutions.
There were just two false alarms produced by
LAV when the default parameters were used. These false alarms were eliminated
when the tool was invoked for the second time with a specified loop
unwinding parameter, and hence there were no false alarms in the
final outputs. In summary, the presented results show that a verification tool like LAV can be
used as a complement to automated testing that improves the
evaluation process.

\begin{table}[h!]
\caption{Summary of bugs in the corpus: the second column
represents the number of students' solutions to the given problem;
the third and the fourth column represents the number of solutions
with bugs detected by manual inspection and by LAV; the fifth column
gives the number of programs shown to be bug-free by LAV (over/under approximation);
the sixth column gives the number of false
alarms made by LAV invoked with default parameters and, if
applicable, with under-approximation.}
\begin{center}
\begin{tabular}{|c|c|c|c|c|c|}
\hline
problem & \# solutions & \# programs   & \# programs & \# bug-free & \# false   \\
        &              & with bugs   & with bugs  &  programs     & alarms with  \\
          &           &by manual&by LAV             &  by LAV             & def./custom      \\
          &           &inspection&             &def./custom            & parameters  \\
\hline
\hline
1.  & 44 & 0  & 0  & 44/-   & 0/- \\ \hline 
2.  & 32 & 11 & 11 & 20/1   & 1/0 \\ \hline 
3.  & 7  & 2  & 2  & -/5    & -/0 \\ \hline 
4.  & 5  & 0  & 1  & 3/1    & 1/0 \\ \hline 
5.  & 12 & 3  & 2  & -/10   & -/0 \\ \hline 
6.  & 7  & 0  & 0  & 6/1    & 1/0 \\ \hline 
7.  & 33 & 0  & 0  & 33/-   & 0/- \\ \hline 
8.  & 31 & 11 & 11 & -/20   & -/0 \\ \hline 
9.  & 10 & 6  & 6  & 4/0    & 0/0 \\ \hline 
10. & 14 & 2  & 2  & 12/0   & 0/0\\ \hline  
11. & 31 & 0  & 0  & 31/-   & 0/- \\ \hline 
12. & 18 & 0  & 0  & 18/-   & 0/- \\ \hline 
13. & 3  & 0  & 0  & 3/-    & 0/- \\ \hline 
14. & 7  & 0  & 0  & 7/-    & 0/- \\ \hline 
15. & 12 & 0  & 0  & 12/-   & 0/- \\ \hline 
\hline
total &266 & 35 &35 &193/38  & 2/0 \\ \hline
\end{tabular}
\end{center}
\label{tab:rezultati}
\end{table}

\subsection{Feedback for students and teachers}

LAV can be used to provide a meaningful and comprehensible feedback
to students while writing their programs. Information like the line number,
the kind of the error, program trace that introduces the error and
values of the variables along this trace, can help student
improve the solution. It can also remind the student to
add an appropriate check that is missing. The example given in Figure
\ref{fig:hint}, extracted from a student's code
written on an exam, shows the error detected by LAV and the generated hint.

\begin{figure}
\begin{verbatim}
                                             verification failed:
1:  #include<stdio.h>                        line 7: UNSAFE
2:  #include<stdlib.h>
3:  int get_digit(int n, int d);             function: main
4:  int main(int argc, char** argv)          error: buffer_overflow
5:  {                                        in line 7: counterexample:
6:  int n, d;                                argc == 1, argv == 1
7:  n = atoi(argv[1]);
8:  d = atoi(argv[2]);                       HINT:
9:  printf("%d\n", get_digit(n, d));         A buffer overflow error occurs when
10: return 0;                                trying to read or write outside the
11: }                                        reserved memory for a buffer/array.
                                             Check the boundaries of the array!
\end{verbatim}
\caption{Listing extracted from student's code written on an
exam (left-hand side) and LAV's  output (right-hand side)}
\label{fig:hint}
\end{figure}

From the software verification support, a teacher can obtain the
information if the student's program contains a bug. The teacher
can use this information in grading assignments by himself. Alternatively,
this information can be taken into account within the wider
integrated framework for obtaining automatically proposed final
grade, as discussed in Section \ref{sec:ag}.

\section{Assignment Evaluation and Structural Similarity of Programs}
\label{sec:similarity}

In this section we propose a similarity measure for programs based on
their control flow graphs, perform its experimental evaluation, and
point to ways it can be used to provide feedback  for students and
teachers.

\subsection{Similarity of CFGs for assignment evaluation}
\label{subsec:simcfg}

To evaluate structural properties of programs, we take the approach
of comparing students' programs to solutions
provided by the teacher. Student's program is considered to be
good if it is similar to some of the programs provided by the teacher
\cite{Wang07}. In order to perform a comparison, a suitable program
representation and a similarity measure are needed. As already noticed
in Section \ref{sec:background}, there is a control flow graph (CFG)
corresponding to each program. The CFG reflects the structure of the
program. Also, there is a linear code sequence attributed to each
node of the CFG which we call the node content. We assume that the
code is in the intermediate LLVM language. In order to measure the
similarity of programs, both the similarity of graphs' structures and
the similarity of node contents
should be considered.
We take the approach of combining the similarity of node contents
with topological similarity of graph nodes described in Section
\ref{sec:background}.

\paragraph{Similarity of node contents.}
The node content is a sequence of LLVM instructions. A simple way of
measuring the similarity of  two sequences of instructions $s_1$ and
$s_2$ is using the edit distance between them $d(s_1,s_2)$ --- the
minimal number of insertion, deletion and substitution operations
over the elements of the sequence by which one sequence can be
transformed into another \cite{Levenshtein66}. In order for edit
distance to be computed, the cost of each
insertion, deletion and substitution operation has to be
defined. We define the cost of
insertion and deletion of an instruction to be 1. Next, we define the
cost of substitution of instruction $i_1$ by instruction $i_2$. Let
$opcode$ be a function that maps an instruction to its opcode (a part
of instruction that specifies the operation to be performed). Let
$opcode(i_1)$ and $opcode(i_2)$ be function calls. Then, the cost of
substitution is 1 if $i_1$ and $i_2$ call different functions, and 0
if they call the same function. If $opcode(i_1)$ or $opcode(i_2)$ is
not a function call, the cost of substitution is 1 if
$opcode(i_1)\neq opcode(i_2)$, and 0 otherwise. Let $n_1=|s_1|$,
$n_2=|s_2|$, and let $M$ be the maximal edit distance over two
sequences of length $n_1$ and $n_2$. Then, the similarity of
sequences $s_1$ and $s_2$ is defined as $1-d(s_1,s_2)/M$.

Although it could be argued that the proposed similarity measure is
rough since it does not  account for differences of instruction
arguments, it is simple, easily implemented, and intuitive.

\paragraph{Full similarity of nodes and similarity of CFGs.}
The topological similarity of nodes can be computed by the method
described in Section \ref{sec:background}. However, purely
topological similarity does not account for differences of the node
content. Hence, we modify the computation of topological similarity
to include the apriori similarity of nodes. The modified update rule is:
$$x^{k+1}_{ij} \leftarrow \sqrt{y_{ij}\cdot\frac{s^{k+1}_{in}(i,j)+s^{k+1}_{out}(i,j)}{2}}$$
where $y_{ij}$ are the similarities of contents of nodes $i$ and $j$
and $s^{k+1}_{in}(i,j)$ and $s^{k+1}_{out}(i,j)$ are defined by
Equations \ref{eq:updaterule}. Also, we set $x^0_{ij}=y_{ij}$. This way,
both content similarity  and topological similarity of nodes are
taken into consideration. The similarity of CFGs can be defined based
on the node similarity matrix as described in Section
\ref{sec:background}. Note that both the similarity of nodes and the
similarity of CFGs take values in the interval $[0,1]$.

It should be noted that our approach provides both the similarity measure for
CFGs and the similarity measure for their nodes ($x_{ij}$). In addition to
evaluating similarity of programs, this approach enables
matching of related parts of the programs by matching
the most similar nodes of CFGs. This could serve as a basis of a
method for suggesting which parts of the student's program could
be further improved.

\subsection{Experimental evaluation}

In order to show that the proposed program similarity measure
corresponds to some intuitive notion of program similarity, we
performed the following experiment. For each program from the corpus
already described in Section \ref{sec:grading}, we found the most
similar program from the rest of the corpus and counted how often
these programs are the solutions for the same problem. That was the
case for 90\% of all programs. This shows that our similarity
measure performs well since with high probability, for each program, the program that is the most similar to
it, corresponds to the same problem. The inspection suggests that in most cases,
where the programs do not correspond to the same problem, student took an
original approach to solving the problem.

The CFGs of the programs from the corpus are rather small.
The average size of CFGs is 15 nodes. The time spent to
compute the similarity of two programs is negligible. However,
out of the educational context where CFGs could have thousands of
nodes, the scalability might be an issue.

\subsection{Feedback for students and teachers}

The students can benefit from program similarity evaluation while
learning and exercising, assuming that the teacher provided a valid
solution or set of solutions to the evaluation system. In
introductory programming courses, most often a student's solution can be considered as better
if it is more similar to one of the
teacher's solutions \cite{Wang07}. In Section $\ref{sec:ag}$
we show that the similarity measure can be used for automatic
calculation of a grade (a feedback that
students easily understand). Moreover, we show that there is a significant
linear dependence of the grade on the similarity value. Due to that
linearity, the similarity value can be considered as an
intuitive feedback, but also it can be translated into descriptive
estimate. For example, the feedback could be that the solution is dissimilar (0-0.5),
roughly similar (0.5-0.7), similar (0.7-0.9) or very similar (0.9-1)
to one of the desired solutions.

The teachers can use the similarity information in automated grading,
as discussed in Section \ref{sec:ag}.

\section{Automated Grading}
\label{sec:ag}

We believe that automated grading can be performed by calculating
a linear combination of different scores measured for the student's
solution. We propose a linear model for prediction of the
teacher-provided grade  of the following form:
$$\hat{y}=\alpha_1 \cdot  x_1 + \alpha_2 \cdot  x_2 + \alpha_3 \cdot  x_3$$
where
\begin{itemize}
\item $\hat{y}$ is the automatically predicted grade,
\item $x_1$ is a result obtained by automated testing
        expressed in the interval $[0,1]$,
\item $x_2$ is 1 if in the student's solution is
        correct as reported by the software verification
        tool, and 0 otherwise,
\item $x_3$ is the maximal value of similarity between
        the student's solution and each of the teacher
        provided solutions (its range is $[0,1]$).
\end{itemize}
It should be noted that we do not use bug count as a parameter, as discussed in
Section \ref{subsec:lavassign}. Different choices for
the coefficients $\alpha_i$, for $i=1,2,3$ could be proposed. In our case, one simple
way could be $\alpha_1=8$, $\alpha_2=1$, and $\alpha_3=1$ since all
programs in our training set won 80\% of the full grade
due to the success in testing. However, it is not always clear
how the teacher's intuitive grading criterion can be factored to
automatically measurable quantities. Teachers need not have the
intuitive feeling for all the variables involved in the grading. For
instance, the behavior of any of the proposed similarity measures
including ours \cite{Wang07,Naude10,Nikolic13} is not clear from
their definitions only. So, it may be unclear how to choose weights
for different variables when combining them in the final grade or if
some of the variables should be nonlinearly transformed in order to
be useful for grading. A natural solution is to try to tune the
coefficients $\alpha_i$, for $i=1,2,3$ so that the behavior of the
predictive model corresponds to the teacher's grading style. For that
purpose, coefficients can be determined automatically using least
squares linear regression \cite{Gross03} if a manually graded corpus
of students' programs is provided by the teacher.

In our evaluation the corpus of programs was split into a training and
a test set where the training set consisted of two thirds of the
corpus and the test set consisted of one third of the corpus. The
training set contained solutions of eight different problems and the
test set contained solutions of remaining seven problems.

Due to the nature of the corpus, for all the instances it holds
$x_1=1$. Therefore, while it is clear that the number of test cases the
program passed ($x_1$) is useful in automated grading, this variable can not
be analyzed based on this corpus.

The optimal values of coefficients $\alpha_i$, $i=1,2,3$, with respect to the
training corpus, are determined using least squares linear
regression. The obtained equation is
$$\hat{y} = 6.058 \cdot  x_1 + 1.014 \cdot  x_2 + 2.919 \cdot  x_3$$
The formula for $\hat{y}$ may seem counterintuitive. Since the minimal
grade in the corpus is 8 and $x_1=1$ for all instances, one would
expect that it holds $\alpha_1\approx 8$. The discrepancy is due to
the fact that for the solutions in the corpus, the minimal value for
$x_3$ is 0.68 --- since the solutions are good (they all passed the
testing) there are no programs with low similarity value. Taking this
into consideration, one can rewrite the formula for $\hat{y}$ as
$$\hat{y} = 8.043 \cdot  x_1 + 1.014 \cdot  x_2 + 0.934 \cdot x'_3$$
where $x'_3 = \frac{x_3-0.68}{1-0.68}$ so the variable $x'_3$ takes
values  from the interval $[0,1]$. This means that when the range of
variability of both $x_2$ and $x_3$ is scaled to the interval $[0,1]$,
their contribution to the mark is rather similar.

Table \ref{tab:corr} shows the comparison between the model
$\hat{y}$ and three other models. Model $\hat{y}_1=8\cdot
x_1+x_2+x_3$ has predetermined parameters, model $\hat{y}_2$ is
trained just with verification information $x_2$ (without similarity
measure), and model $\hat{y}_3$ is trained only with similarity
measure $x_3$ (without verification information). Results show
that the performance of model $\hat{y}$ on the test set
(consisting of problems not appearing in the training set) is
outstanding --- the correlation is $0.842$ and the model accounts for
$71\%$ of the variability of teacher provided grade. These results
indicate a strong and reliable dependence between teacher provided grade and the
variables $x_i$, meaning that a grade can be reliably predicted by $\hat{y}$.
Also, $\hat{y}$ is much better than other models. This shows that the approach using both
verification information and graph similarity information is superior to
approaches using only one source of information, and also that
automated tuning of coefficients of the model provides better
prediction than giving them in advance.

Inspection of solutions that yielded the biggest error in
prediction suggests that the greatest source of discrepancy of
automatically provided and teacher provided grades are the original
solutions given by students and the solutions that the teacher did
not predict in advance. However, we cannot exclude other factors apart form presence
of bugs and similarity to model solutions, that govern human grading
process.

\begin{table}
\begin{center}
\begin{tabular}{lccc}
\hline
\hspace{8mm} & \hspace{2mm}$r$\hspace{2mm} & \hspace{2mm}$r^2\cdot 100\%$\hspace{2mm} & \hspace{2mm}Rel. error\hspace{2mm}\\
\hline
$\hat{y}$ & 0.842 & 71\% & 10.1\%\\
$\hat{y}_1$  & 0.730 & 53.3\% & 12.8\%\\
$\hat{y}_2$ & 0.620 & 38.4\% & 16.7\%\\
$\hat{y}_3$ & 0.457 & 20.9\% & 17.7\%\\
\hline
\end{tabular}
\end{center}
\caption{The performance of the predictive model on the training and
test set. We provide correlation  coefficient ($r$), the fraction of
variance of $y$ accounted by the model ($100\cdot r^2$), and relative
error --- average error divided by the length of the range in which
the grades vary (which is 8 to 10 in the case of this particular
corpus).} \label{tab:corr}
\end{table}

\section{Related work}
\label{sec:rw}
Automated testing is the most common way of evaluating students'
programs \cite{douce05}. Test cases are usually supplied by a teacher
and/or randomly generated \cite{MandalMR07}. A
lot of systems use this approach, for example, PSGE \cite{psge},
Kassandra \cite{kassandra}, BOSS \cite{boss}, WebToTeach
\cite{webtoteach}, Schemerobe \cite{scheme}, TRY  \cite{try}, HoGG
\cite{hogg}, BAGS \cite{bags}, on-line Judge \cite{judge}, JEWL
\cite{jewl}, Quiver \cite{quiver}, and JUnit \cite{junit}.

Software verification techniques are not commonly used in automated
evaluation of programs. There are limited experiments on using Java PathFinder
model checker for automated test case generation \cite{ihantola07}.
Tools with integrated support for automated testing and verification,
e.g. Ceasar \cite{garavel98}, are usually too complex and not aimed
for educational purposes. To the authors' knowledge, there is no other
software verification tool deployed in process of automated bug
finding as a complement to automated testing of students' programs.
The tool LAV was already used, to a limited extent, for finding bugs in students' programs \cite{vstteLAV12}. In that work, a different sort of  corpus was used, as discussed in Section \ref{subsec:lavassign}.
Also, that application did not aim at automated grading, and instead was made in the wider context of design and development of LAV as a general-purpose SMT-based error finding platform.

Wang et al. proposed a grading approach for assignments in C based
only on program similarity \cite{Wang07}. It relies on dependence
graphs \cite{horwitz} as program representation. They perform various code
transformations in order to standardize the representation of the
program. In this approach, the similarity is
calculated based on comparison of structure, statement, and size
which are weighted by some predetermined coefficients.
Their approach is evaluated on  10 problems, 200 solutions each, and obtain good
results compared to manual grading. Manual grading was performed strictly according to the
criterion that indicates how the scores are awarded for structure,
statements used, and size. However, it is not quite obvious that human grading
is always expressed strictly in terms of these three factors.
An advantage of our approach compared to this one is automated tuning of
weights corresponding to different variables used in grading, instead of using the
predetermined ones. Since teachers do not need to have an intuitive feeling
for different similarity measures, it may be unclear how the
corresponding weights should be chosen. Also, we avoid language
dependent transformations by using LLVM which makes our approach applicable
to large variety of programming languages. Very similar approach to the
one of Wang et al. was presented by Li et al.
\cite{Li10}.

Another approach to grading assignments based only on graph
similarity measure is proposed by Naud\'e et al. \cite{Naude10}. They
represent programs as dependence graphs and propose directed acyclic
graph (DAG) similarity measure.  In
their approach, for each solution to be graded, several similar
solutions in the training set are found and the grade is formed by
combining grades of these solutions with respect to matched portions
of the similar solutions. The approach was evaluated on one
assignment problem and the correlation between human and machine provided
grades is the same as ours. For appropriate grading they recommend at least 20 manually
graded solutions of various qualities for each problem to be automatically graded.
In the case of automatic grading of high quality solutions
(as is the case with our corpus), using 20 manually graded
solutions, their approach achieves 16.7\% relative error, while with
90 manually graded solutions it achieves around 10\%.
The improvement that our approach provides is reflected through several indicators.
We used a heterogeneous corpus of 15 problems instead of one.
Our approach uses 1 to 3 model solutions for each problem to be
graded and a training set for weight estimation which does not need
to contain the solutions for the program to be graded.
So, after the initial training has been performed, for
each new problem only few model solutions should be provided. Using
1 to 3 model solutions, we achieve 10\% relative error (see Table
\ref{tab:corr}). Due to the use of the LLVM platform, we do not use
language dependent transformations, so our approach is
applicable to large number of programming languages.
The similarity measure we use, called neighbor matching, is similar to the one of
Naud\'e et al., but for our measure, important theoretical properties (e.g. convergence)
are proven \cite{Nikolic13}. The neighbor matching method was already applied to several
problems but in all these applications its use was limited to ordinary graphs with nodes without
any internal specifics. In order to be applied to CFGs, the method was modified to
include node content similarity which was independently defined as described in Section \ref{subsec:simcfg}.

Finally, as a distinctive feature of our system, we are not aware of
open source implementations of the similarity based approaches.
A drawback in the comparison of our approach to previously described ones is that
our corpus consists of high quality solutions due to the grading setting
at the course.

Apart of assignment grading, regression techniques were also used for
final grade forecasting with good results. For this purpose, Macfadyen et al. used data from learning management system and identified variables most useful for the prediction, e.g., number of assessments completed and number of discussion and mail messages sent \cite{Macfadyen10}.
Kotsiantis performed successful forecasting based on demographic characteristics of students, results of several written assignments, and class attendance \cite{Kotsiantis12}.

\section{Conclusions and Further Work}
\label{sec:conclusions}

We presented two techniques that can be used
for improving automated evaluation of students' programs. First
one is based on software verification and second one on CFG
similarity measurement. Both techniques can be used for providing
useful and helpful feedback to students and for improving automated
grading for teachers. In our evaluation, we show that synergy of
these techniques offers more information useful for
automated grading than any of them independently. Also, we obtained
good results in prediction of the grades for a new set of assignments. This
shows that our approach can be trained to adapt to teacher's
grading style on several teacher graded problems and then be used on
different problems using only few model solutions per
problem. An important advantage of our approach is independence of
specific programming language since LLVM platform (which we use to
produce intermediate code) supports large number of
programming languages. We also provide the corresponding open source
tools.

In our future work we are planning to make an integrated web-based
system with support for the
mentioned techniques along with compiling, automated testing,
profiling and detection of plagiarism of students' programs. Also, we
intend to improve feedback to students by indicating missing or redundant
parts of code compared to the teacher's solution. This feature would
rely on the fact that our similarity measure provides the similarity
values for nodes of CFGs, and hence enables matching the parts of code
between two solutions. If some parts of
the solutions cannot be matched or are matched with very low
similarity, this can be reported to the student. On the other hand,
the similarity of the CFG with itself could reveal the repetitions of
parts of the code and suggest that refactoring could be performed.
We are planning to integrate LLVM-based open source tool KLEE
\cite{KLEE} for automated test case generation and also to add
support for teacher supplied test cases.

We are also planning to explore potential for using software
verification tools for proving functional correctness of students'
programs. This task would pose new challenges. Testing,
profiling, bug finding and similarity measurement are used on original
students' programs, which makes the automation easy.
For verification of functional correctness, the teacher
would have to define correctness conditions (possibly in terms of
implemented functions) and insert corresponding assertions in appropriate places in students'
programs which should be possible to
automate in some cases, but it is not trivial in general. In addition,
for some programs it is not easy to formulate correctness conditions
(for example, for programs that are expected only to print some
messages on standard output).

\bibliographystyle{theapa}

\begin{thebibliography}{}

\bibitem[\protect\BCAY{{Ala-Mutka}}{{Ala-Mutka}}{2005}]{mutka05}
{Ala-Mutka}, K.~M. \BBOP2005\BBCP.
\newblock \BBOQ {A Survey of Automated Assessment Approaches for Programming
  Assignments}\BBCQ\
\newblock {\Bem Computer Science Education}, {\Bem 15}, 83--102.

\bibitem[\protect\BCAY{Allen\ \BBA\ Seaman}{Allen\ \BBA\
  Seaman}{2010}]{Allen09}
Allen, I.~E.\BBACOMMA\  \BBA\ Seaman, J. \BBOP2010\BBCP.
\newblock \BBOQ Learning on demand: Online education in the united states,
  2009\BBCQ\
\newblock \BTR, The Sloan Consortium.

\bibitem[\protect\BCAY{Arnow\ \BBA\ Barshay}{Arnow\ \BBA\
  Barshay}{1999}]{webtoteach}
Arnow, D.\BBACOMMA\  \BBA\ Barshay, O. \BBOP1999\BBCP.
\newblock \BBOQ Webtoteach: an interactive focused programming exercise
  system\BBCQ\
\newblock {\Bem Frontiers in Education, Annual}, {\Bem 1},
  12A9/39--12A9/44vol.1.

\bibitem[\protect\BCAY{Barrett, Sebastiani, Seshia,\ \BBA\ Tinelli}{Barrett
  et~al.}{2009}]{BSST09HBSAT}
Barrett, C., Sebastiani, R., Seshia, S.~A., \BBA\ Tinelli, C. \BBOP2009\BBCP.
\newblock \BBOQ Satisfiability modulo theories\BBCQ\
\newblock In {\Bem Handbook of Satisfiability}, \lowercase{\BVOL}\ 185 of {\Bem
  Frontiers in Artificial Intelligence and Applications}, \BPGS\ 825--885. IOS
  Press.

\bibitem[\protect\BCAY{Blondel, Gajardo, Heymans, Snellart,\ \BBA\ van
  Dooren}{Blondel et~al.}{2004}]{Blondel04}
Blondel, V.~D., Gajardo, A., Heymans, M., Snellart, P., \BBA\ van Dooren, P.
  \BBOP2004\BBCP.
\newblock \BBOQ A measure of similarity between graph vertices: Applications to
  synonym extraction and web searching\BBCQ\
\newblock {\Bem SIAM Review}, {\Bem 46}, 647---666.

\bibitem[\protect\BCAY{Cadar, Dunbar,\ \BBA\ Engler}{Cadar et~al.}{2008}]{KLEE}
Cadar, C., Dunbar, D., \BBA\ Engler, D. \BBOP2008\BBCP.
\newblock \BBOQ Klee: Unassisted and automatic generation of high-coverage
  tests for complex systems programs\BBCQ\
\newblock In {\Bem Proceeding OSDI'08 Proceedings of the 8th USENIX conference
  on Operating systems design and implementation}. USENIX Association Berkeley.

\bibitem[\protect\BCAY{Cheang, Kurnia, Lim,\ \BBA\ Oon}{Cheang
  et~al.}{2003}]{judge}
Cheang, B., Kurnia, A., Lim, A., \BBA\ Oon, W.-C. \BBOP2003\BBCP.
\newblock \BBOQ On automated grading of programming assignments in an academic
  institution\BBCQ\
\newblock {\Bem Comput. Educ.}, {\Bem 41\/}(2), 121--131.

\bibitem[\protect\BCAY{Chipounov, Kuznetsov,\ \BBA\ Candea}{Chipounov
  et~al.}{2011}]{S2E}
Chipounov, V., Kuznetsov, V., \BBA\ Candea, G. \BBOP2011\BBCP.
\newblock \BBOQ S2e: a platform for in-vivo multi-path analysis of software
  systems\BBCQ\
\newblock {\Bem SIGARCH Comput. Archit. News}, {\Bem 39}, 265--278.

\bibitem[\protect\BCAY{Clarke, Kroening,\ \BBA\ Lerda}{Clarke
  et~al.}{2004}]{CBMC}
Clarke, E., Kroening, D., \BBA\ Lerda, F. \BBOP2004\BBCP.
\newblock \BBOQ A tool for checking ansi-c programs\BBCQ\
\newblock In {\Bem In Tools and Algorithms for the Construction and Analysis of
  Systems}, \BPGS\ 168--176. Springer.

\bibitem[\protect\BCAY{Clarke}{Clarke}{2008}]{modelc}
Clarke, E.~M. \BBOP2008\BBCP.
\newblock {\Bem 25 Years of Model Checking --- The Birth of Model Checking},
  \lowercase{\BVOL}\ 5000/2008, 1--26 of {\Bem Lecture Notes in Computer
  Science}.
\newblock Springer.

\bibitem[\protect\BCAY{Cordeiro, Fischer,\ \BBA\ Marques-Silva}{Cordeiro
  et~al.}{2009}]{ESBMC}
Cordeiro, L., Fischer, B., \BBA\ Marques-Silva, J. \BBOP2009\BBCP.
\newblock \BBOQ Smt-based bounded model checking for embedded ansi-c
  software\BBCQ\
\newblock {\Bem International Conference on Automated Software Engineering},
  {\Bem 0}, 137--148.

\bibitem[\protect\BCAY{Cousot\ \BBA\ Cousot}{Cousot\ \BBA\
  Cousot}{1977}]{cousot77}
Cousot, P.\BBACOMMA\  \BBA\ Cousot, R. \BBOP1977\BBCP.
\newblock \BBOQ Abstract interpretation: A unified lattice model for static
  analysis of programs by construction or approximation of fixpoints\BBCQ\
\newblock In {\Bem POPL}, \BPGS\ 238--252.

\bibitem[\protect\BCAY{Douce, Livingstone,\ \BBA\ Orwell}{Douce
  et~al.}{2005}]{douce05}
Douce, C., Livingstone, D., \BBA\ Orwell, J. \BBOP2005\BBCP.
\newblock \BBOQ {Automatic test-based assessment of programming: A
  review}\BBCQ\
\newblock {\Bem J. Educ. Resour. Comput.}, {\Bem 5\/}(3), 4+.

\bibitem[\protect\BCAY{Ellsworth, Fenwick,\ \BBA\ Kurtz}{Ellsworth
  et~al.}{2004}]{quiver}
Ellsworth, C.~C., Fenwick, Jr., J.~B., \BBA\ Kurtz, B.~L. \BBOP2004\BBCP.
\newblock \BBOQ The quiver system\BBCQ\
\newblock In {\Bem Proceedings of the 35th SIGCSE technical symposium on
  Computer science education}, SIGCSE '04, \BPGS\ 205--209, New York, NY, USA.
  ACM.

\bibitem[\protect\BCAY{English}{English}{2004}]{jewl}
English, J. \BBOP2004\BBCP.
\newblock \BBOQ Automated assessment of gui programs using jewl\BBCQ\
\newblock {\Bem SIGCSE Bull.}, {\Bem 36\/}(3), 137--141.

\bibitem[\protect\BCAY{Garavel}{Garavel}{1998}]{garavel98}
Garavel, H. \BBOP1998\BBCP.
\newblock \BBOQ Open/c{\ae}sar: An open software architecture for verification,
  simulation, and testing\BBCQ\
\newblock In Steffen, B.\BED, {\Bem Proceedings of the First International
  Conference on Tools and Algorithms for the Construction and Analysis of
  Systems TACAS'98 (Lisbon, Portugal)}, \lowercase{\BVOL}\ 1384 of {\Bem
  Lecture Notes in Computer Science}, \BPGS\ 68--84, Berlin. Springer Verlag.
\newblock Full version available as INRIA Research Report~RR-3352.

\bibitem[\protect\BCAY{Godefroid, Levin,\ \BBA\ Molnar}{Godefroid
  et~al.}{2012}]{godefroid12}
Godefroid, P., Levin, M.~Y., \BBA\ Molnar, D.~A. \BBOP2012\BBCP.
\newblock \BBOQ Sage: Whitebox fuzzing for security testing\BBCQ\
\newblock {\Bem ACM Queue}, {\Bem 10\/}(1), 20.

\bibitem[\protect\BCAY{Gross}{Gross}{2003}]{Gross03}
Gross, J. \BBOP2003\BBCP.
\newblock {\Bem Linear Regression}.
\newblock Springer.

\bibitem[\protect\BCAY{Hext\ \BBA\ Winings}{Hext\ \BBA\ Winings}{1969}]{psge}
Hext, J.~B.\BBACOMMA\  \BBA\ Winings, J.~W. \BBOP1969\BBCP.
\newblock \BBOQ An automatic grading scheme for simple programming
  exercises\BBCQ\
\newblock {\Bem Commun. ACM}, {\Bem 12\/}(5), 272--275.

\bibitem[\protect\BCAY{Heymans\ \BBA\ Singh}{Heymans\ \BBA\
  Singh}{2003}]{Heymans03}
Heymans, M.\BBACOMMA\  \BBA\ Singh, A. \BBOP2003\BBCP.
\newblock \BBOQ Deriving phylogenetic trees from the similarity analysis of
  metabolic pathways\BBCQ\
\newblock {\Bem Bioinformatics}, {\Bem 19}, 138---146.

\bibitem[\protect\BCAY{Horwitz\ \BBA\ Reps}{Horwitz\ \BBA\
  Reps}{1992}]{horwitz}
Horwitz, S.\BBACOMMA\  \BBA\ Reps, T. \BBOP1992\BBCP.
\newblock \BBOQ The use of program dependence graphs in software
  engineering\BBCQ\
\newblock In {\Bem Proceedings of the 14th international conference on Software
  engineering}, ICSE '92, \BPGS\ 392--411, New York, NY, USA. ACM.

\bibitem[\protect\BCAY{Ihantola}{Ihantola}{2007}]{ihantola07}
Ihantola, P. \BBOP2007\BBCP.
\newblock \BBOQ Creating and visualizing test data from programming
  exercises\BBCQ\
\newblock {\Bem Informatics in education}, {\Bem 6\/}(1), 81--102.

\bibitem[\protect\BCAY{Ihantola, Ahoniemi, Karavirta,\ \BBA\
  Sepp\"{a}l\"{a}}{Ihantola et~al.}{2010}]{ihantola10}
Ihantola, P., Ahoniemi, T., Karavirta, V., \BBA\ Sepp\"{a}l\"{a}, O.
  \BBOP2010\BBCP.
\newblock \BBOQ Review of recent systems for automatic assessment of
  programming assignments\BBCQ\
\newblock In {\Bem Proceedings of the 10th Koli Calling International
  Conference on Computing Education Research}, Koli Calling '10, \BPGS\ 86--93,
  New York, NY, USA. ACM.

\bibitem[\protect\BCAY{Jones}{Jones}{2001}]{try}
Jones, E.~L. \BBOP2001\BBCP.
\newblock \BBOQ Grading student programs - a software testing approach\BBCQ\
\newblock {\Bem Journal of Computing Sciences in Colleges}, {\Bem 16 (2)},
  187--194.

\bibitem[\protect\BCAY{Joy, Griffiths,\ \BBA\ Boyatt}{Joy et~al.}{2005}]{boss}
Joy, M., Griffiths, N., \BBA\ Boyatt, R. \BBOP2005\BBCP.
\newblock \BBOQ The boss online submission and assessment system\BBCQ\
\newblock {\Bem J. Educ. Resour. Comput.}, {\Bem 5\/}(3).

\bibitem[\protect\BCAY{King}{King}{1976}]{King76}
King, J.~C. \BBOP1976\BBCP.
\newblock \BBOQ Symbolic execution and program testing\BBCQ\
\newblock {\Bem Communications of the ACM}, {\Bem 19\/}(7), 385--394.

\bibitem[\protect\BCAY{Kleinberg}{Kleinberg}{1999}]{Kleinberg99}
Kleinberg, J.~M. \BBOP1999\BBCP.
\newblock \BBOQ Authoritative sources in a hyperlinked environment\BBCQ\
\newblock {\Bem Journal of the ACM}, {\Bem 46}, 604 --- 632.

\bibitem[\protect\BCAY{Kotsiantis}{Kotsiantis}{2012}]{Kotsiantis12}
Kotsiantis, S.~B. \BBOP2012\BBCP.
\newblock \BBOQ Use of machine learning techniques for educational proposes: a
  decision support system for forecasting students' grades\BBCQ\
\newblock {\Bem Artificial Intelligence Review}, {\Bem 34\/}(4), 331--344.

\bibitem[\protect\BCAY{Kuhn}{Kuhn}{1955}]{Kuhn55}
Kuhn, H.~W. \BBOP1955\BBCP.
\newblock \BBOQ The hungarian method for the assignment problem\BBCQ\
\newblock {\Bem Naval Research Logistics Quarterly}, {\Bem 2\/}(1-2), 83--97.

\bibitem[\protect\BCAY{Lattner\ \BBA\ Adve}{Lattner\ \BBA\ Adve}{2002}]{llvm}
Lattner, C.\BBACOMMA\  \BBA\ Adve, V. \BBOP2002\BBCP.
\newblock \BBOQ {The LLVM Instruction Set and Compilation Strategy}\BBCQ.

\bibitem[\protect\BCAY{Levenshtein}{Levenshtein}{1966}]{Levenshtein66}
Levenshtein, V.~I. \BBOP1966\BBCP.
\newblock \BBOQ Binary codes capable of correcting deletions, insertions, and
  reversals\BBCQ\
\newblock {\Bem Soviet Physics Doklady}, {\Bem 10\/}(8), 707--710.

\bibitem[\protect\BCAY{Li, Pan, Zhang, Chen, Nie,\ \BBA\ He}{Li
  et~al.}{2010}]{Li10}
Li, J., Pan, W., Zhang, R., Chen, F., Nie, S., \BBA\ He, X. \BBOP2010\BBCP.
\newblock \BBOQ Design and implementation of semantic matching based automatic
  scoring system for c programming language\BBCQ\
\newblock In {\Bem Proceedings of the Entertainment for education, and 5th
  international conference on E-learning and games}, \BPGS\ 247--257.
  Springer-Verlag.

\bibitem[\protect\BCAY{Macfadyen\ \BBA\ Dawson}{Macfadyen\ \BBA\
  Dawson}{2010}]{Macfadyen10}
Macfadyen, L.~P.\BBACOMMA\  \BBA\ Dawson, S. \BBOP2010\BBCP.
\newblock \BBOQ Mining lms data to develop an "early warning system" for
  educators: a proof of concept\BBCQ\
\newblock {\Bem Computers and Education}, {\Bem 54\/}(2), 588--599.

\bibitem[\protect\BCAY{Mandal, Mandal,\ \BBA\ Reade}{Mandal
  et~al.}{2007}]{MandalMR07}
Mandal, A.~K., Mandal, C.~A., \BBA\ Reade, C. \BBOP2007\BBCP.
\newblock \BBOQ A system for automatic evaluation of c programs: Features and
  interfaces\BBCQ\
\newblock {\Bem IJWLTT}, {\Bem 2\/}(4), 24--39.

\bibitem[\protect\BCAY{Matt}{Matt}{1994}]{kassandra}
Matt, U.~V. \BBOP1994\BBCP.
\newblock \BBOQ Kassandra: The automatic grading system\BBCQ\
\newblock {\Bem SIGCUE Outlook}, {\Bem 22}, 22--26.

\bibitem[\protect\BCAY{Morris}{Morris}{2002}]{hogg}
Morris, D.~S. \BBOP2002\BBCP.
\newblock \BBOQ Automatically grading java programming assignments via
  reflection, inheritance, and regular expressions\BBCQ\
\newblock {\Bem Frontiers in Education Conference 1}, {\Bem 1}, T3G--22.

\bibitem[\protect\BCAY{Morris}{Morris}{2003}]{bags}
Morris, D. \BBOP2003\BBCP.
\newblock \BBOQ Automatic grading of student's programming assignments: an
  interactive process and suit of programs\BBCQ\
\newblock In {\Bem Proceedings of the Frontiers in Education Conference 3},
  \lowercase{\BVOL}~3, \BPGS\ 1--6.

\bibitem[\protect\BCAY{Naud{\'e}, Greyling,\ \BBA\ Vogts}{Naud{\'e}
  et~al.}{2010}]{Naude10}
Naud{\'e}, K.~A., Greyling, J.~H., \BBA\ Vogts, D. \BBOP2010\BBCP.
\newblock \BBOQ Marking student programs using graph similarity\BBCQ\
\newblock {\Bem Computers and Education}, {\Bem 54\/}(2), 545--561.

\bibitem[\protect\BCAY{Nikoli\'c}{Nikoli\'c}{2013}]{Nikolic13}
Nikoli\'c, M. \BBOP2013\BBCP.
\newblock \BBOQ Measuring similarity of graph nodes by neighbor matching\BBCQ\
\newblock {\Bem Intelligent Data Analysis}, {\Bem Accepted for publication}.

\bibitem[\protect\BCAY{Nipkow}{Nipkow}{2012}]{nipkow}
Nipkow, T. \BBOP2012\BBCP.
\newblock \BBOQ Teaching semantics with a proof assistant: No more lsd trip
  proofs\BBCQ\
\newblock In {\Bem VMCAI}, \BPGS\ 24--38.

\bibitem[\protect\BCAY{Pears, Seidman, Malmi, Mannila, Adams, Bennedsen,
  Devlin,\ \BBA\ Paterson}{Pears et~al.}{2007}]{Pears07}
Pears, A., Seidman, S., Malmi, L., Mannila, L., Adams, E., Bennedsen, J.,
  Devlin, M., \BBA\ Paterson, J. \BBOP2007\BBCP.
\newblock \BBOQ A survey of literature on the teaching of introductory
  programming\BBCQ\
\newblock In {\Bem Working group reports on ITiCSE on Innovation and technology
  in computer science education}, ITiCSE-WGR '07, \BPGS\ 204--223, New York,
  NY, USA. ACM.

\bibitem[\protect\BCAY{Saikkonen, Malmi,\ \BBA\ Korhonen}{Saikkonen
  et~al.}{2001}]{scheme}
Saikkonen, R., Malmi, L., \BBA\ Korhonen, A. \BBOP2001\BBCP.
\newblock \BBOQ {Fully automatic assessment of programming exercises}\BBCQ\
\newblock {\Bem ACM Sigcse Bulletin}, {\Bem 33}, 133--136.

\bibitem[\protect\BCAY{Tassey}{Tassey}{2002}]{tassey02economic}
Tassey, G. \BBOP2002\BBCP.
\newblock \BBOQ {The economic impacts of inadequate infrastructure for software
  testing}\BBCQ\
\newblock \BTR, National Institute of Standards and Technology.

\bibitem[\protect\BCAY{Tillmann\ \BBA\ Halleux}{Tillmann\ \BBA\
  Halleux}{2008}]{PEX}
Tillmann, N.\BBACOMMA\  \BBA\ Halleux, J. \BBOP2008\BBCP.
\newblock \BBOQ Pex – white box test generation for .net\BBCQ\
\newblock In {\Bem Proc. of TAP 2008, the 2nd International Conference on Tests
  and Proofs}, \lowercase{\BVOL}\ 4966 of {\Bem LNCS}, \BPGS\ 134--153.
  Springer.

\bibitem[\protect\BCAY{Vujo\v{s}evi\'{c}-Jani\v{c}i\'{c}\ \BBA\
  Kuncak}{Vujo\v{s}evi\'{c}-Jani\v{c}i\'{c}\ \BBA\ Kuncak}{2012}]{vstteLAV12}
Vujo\v{s}evi\'{c}-Jani\v{c}i\'{c}, M.\BBACOMMA\  \BBA\ Kuncak, V.
  \BBOP2012\BBCP.
\newblock \BBOQ Development and evaluation of {LAV}: an {SMT}-based error
  finding platform\BBCQ\
\newblock In {\Bem Verified Software: Theories, Tools and Experiments (VSTTE)},
  LNCS.

\bibitem[\protect\BCAY{Vujo\v{s}evi\'c-Jani\v{c}i\'c\ \BBA\
  To\v{s}i\'c}{Vujo\v{s}evi\'c-Jani\v{c}i\'c\ \BBA\
  To\v{s}i\'c}{2008}]{mvjtosic}
Vujo\v{s}evi\'c-Jani\v{c}i\'c, M.\BBACOMMA\  \BBA\ To\v{s}i\'c, D.
  \BBOP2008\BBCP.
\newblock \BBOQ The role of programming paradigms in the first programming
  courses\BBCQ\
\newblock {\Bem The Teaching of Mathematics}, {\Bem XI\/}(2), 63--83.

\bibitem[\protect\BCAY{Wang, Su, Wang,\ \BBA\ Ma}{Wang et~al.}{2007}]{Wang07}
Wang, T., Su, X., Wang, Y., \BBA\ Ma, P. \BBOP2007\BBCP.
\newblock \BBOQ Semantic similarity-based grading of student programs\BBCQ\
\newblock {\Bem Information and Software Technology}, {\Bem 49\/}(2), 99--107.

\bibitem[\protect\BCAY{Wick, Stevenson,\ \BBA\ Wagner}{Wick
  et~al.}{2005}]{junit}
Wick, M., Stevenson, D., \BBA\ Wagner, P. \BBOP2005\BBCP.
\newblock \BBOQ Using testing and junit across the curriculum\BBCQ\
\newblock {\Bem SIGCSE Bull.}, {\Bem 37\/}(1), 236--240.

\end{thebibliography}

\end{document}